%% file: main.tex
\documentclass{article}

\PassOptionsToPackage{numbers,sort&compress}{natbib}
\usepackage[main, final]{neurips_2026}

\usepackage[utf8]{inputenc}
\usepackage[T1]{fontenc}
\usepackage{hyperref}
\usepackage{url}
\usepackage{booktabs}
\usepackage{amsfonts}
\usepackage{amsmath}
\usepackage{amssymb}
\usepackage{nicefrac}
\usepackage{microtype}
\usepackage{xcolor}
\usepackage{graphicx}
\usepackage{multirow}
\usepackage{algorithm}
\usepackage{algpseudocode}
\usepackage[normalem]{ulem}
\usepackage{makecell}
\usepackage{wrapfig}
\usepackage{caption}
\newcommand{\method}{BiLoCo}

\newtheorem{theorem}{Theorem}[section]

\newtheorem{proposition}[theorem]{Proposition}

\title{\method{}: Binary Low-Rank Corrections for \\  Post-Training Quantization}

\author{
{David Jin\thanks{Equal contribution.} \quad
 Beshr IslamBouli\footnotemark[1] \quad
 Tarushii Goel \quad
 Han Guo \quad
 Yoon Kim}
\vspace{2mm} \\ 
Massachusetts Institute of Technology \vspace{2mm} \\ 
\texttt{\{jindavid,beshr,tarushii,hanguo,yoonkim\}@mit.edu}
}

\begin{document}

\maketitle

\input{sections/00_abstract}
\input{sections/01_introduction}
\input{sections/02_background}
\input{sections/03a_method}
\input{sections/03b_theory}
\input{sections/03c_kernel}

\input{sections/05_experiments}
\input{sections/06_related_work}
\input{sections/07_limitations}
\input{sections/08_conclusion}

\small
\bibliographystyle{plainnat}
\bibliography{sections/references}

\clearpage
\appendix
\input{sections/a2_additional_experiments}
\input{sections/a1_proofs}

\end{document}

%% file: sections/00_abstract.tex
\begin{abstract} 

Post-training quantization  (PTQ) to FP4 has emerged as a key technique for reducing large language model inference cost, especially on Blackwell GPUs, which offer native FP4 tensor core support. 
Direct PTQ to  FP4, however, still leaves a noticeable accuracy gap. 
Recent work on SVDQuant~\cite{li2025svdquant} mitigates the errors introduced through FP4 quantization for diffusion models by adding high-precision low-rank corrections obtained via SVD. 
However, this high-precision low-rank correction requires many bits per rank, and it runs on the same tensor cores as the main FP4 GEMM.
We introduce \method{}, a low-rank correction method whose left and right factors are stored as binary ${\{\pm 1\}}$ sign vectors rather than as BF16 numbers. 
The signed format uses $16{\times}$ less storage per rank-one component, and since the correction now consists of additions/subtractions, this can be run on CUDA cores concurrently with the FP4 tensor-core GEMM. 
On 4B--72B LLMs, \method{} matches or improves memory-matched SVDQuant for PTQ. 
We implement \method{} on B300 as a fused single kernel, where the FP4 GEMM runs on tensor cores, the bit-packed correction on the CUDA cores of the same launch, and the two are combined in the GEMM epilogue. 
With batch size $1$, the fused kernel is at most $11\%$ slower than the FP4-only GEMM for every rank up to $512$.
At rank $128$, it is at most $20\%$ slower up to batch size $8$.
\end{abstract}

%% file: sections/01_introduction.tex
\section{Introduction}
\label{sec:intro}

LLM decode is dominated by weight movement~\cite{guo-etal-2024-fast, lin2025qservewakv, 10.1145/3710848.3710871}. 
Post-training quantization (PTQ) to FP4 on Blackwell GPUs can thus improve latency by reducing this traffic and moreover exposing native FP4 tensor-core GEMM. 
However, direct post-training quantization to FP4 still leaves an accuracy gap. 
A natural fix is to keep the residual weight matrix in lower bits and add a low-rank higher-precision correction, i.e., factorize the original 16-bit matrix ${W}$  as ${W} \approx  R + VU$ where $R$ is  a 4 bit matrix and $VU$ is kept in higher precision. 
Such a correction approach was investigated for weight-only quantization in LQ-LoRA~\cite{guo2024lqloralowrankplusquantized}, LoftQ~\cite{li2023loftq}, and CALDERA~\cite{saha2024compressinglargelanguagemodels}. SVDQuant~\cite{li2025svdquant} extends this approach to weight and activation quantization to FP4 with a BF16 rank-$32$ correction, where the correction absorbs weight outliers that the low-bit residual cannot represent. 

The initial SVDQuant work focused on vision diffusion models. However, autoregressive LLM decode regime differs from the diffusion regime in ways that affect correction design. 
Concretely, with LLM decode shapes, the FP4 GEMM is weight-traffic bound, causing tensor cores to stall on HBM while CUDA cores sit idle. 
Tensor cores and CUDA cores are separate units that issue instructions independently, and recent attention kernels overlap work on the two~\cite{shah2024flashattention3fastaccurateattention,zadouri2026flashattention4algorithmkernelpipelining} (\S\ref{sec:background_gpu}).
A correction that uses CUDA cores rather than tensor cores could therefore run concurrently with the FP4 GEMM instead of competing for the same hardware.
SVDQuant, however, computes its BF16 correction on tensor cores, the same units that run the FP4 GEMM.
The challenge then, is to find a correction format whose arithmetic naturally fits the CUDA-core slot.

\begin{figure}[t]
\centering
\includegraphics[width=0.96\textwidth]{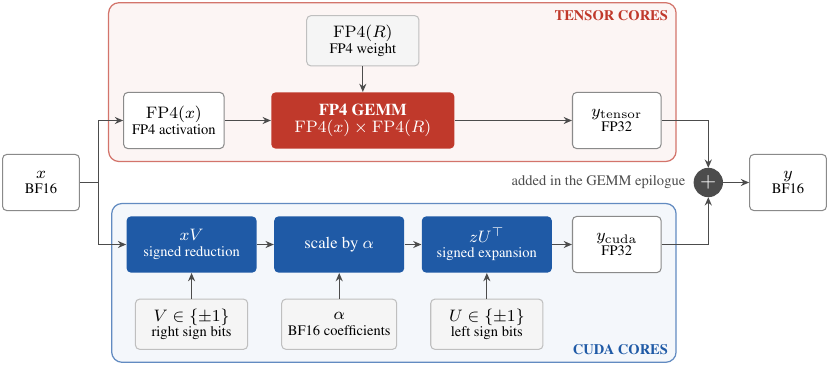}
\caption{\method{} inference for performing $y = xR + xVU^\top.$ A native-FP4 residual GEMM on tensor
cores runs alongside a signed add/subtract correction on CUDA cores.}
\vspace{-2mm}
\label{fig:figure1}\vspace{-5mm}
\end{figure}

Could we use an alternative factorization of the weight matrix to achieve strong error correction while ensuring the correction component can make use of idle CUDA cores during inference? 
We introduce Binary Low-rank Correction (\method{}, Figure~\ref{fig:figure1}), where  instead of storing BF16 left and right singular vectors, we store the correction factors as $\{\pm 1\}$ sign vectors, together with scalar coefficients. 
Each rank-one component then reduces to signed accumulation, replacing multiply operations with predicated add/subtract operations on CUDA cores. 
The representation also changes the storage trade-off: ignoring the per-term scalar coefficient, one binary vector pair costs $16\times$ fewer bits than a BF16 vector pair, so the same vector-storage budget as a BF16 rank-$k$ branch can store a binary rank-$16k$ branch.

The main algorithmic challenge is that the best sign-vector decomposition is not given by the SVD. 
Following \citet{riasanovsky2024efficient1bittensorapproximations}, we fit the sign vectors one rank-one component at a time.
For each component, we start from random signs, alternate between updating $u$ and $v$ for a few steps, compute the best scale, and subtract the component from the residual.
After all components are fit, we re-solve all scales jointly by least squares, which accounts for the overlap between components, and quantize the remaining residual to NVFP4.

We find that across various LLM families, \method{} is effective at the same storage budget as the BF16 correction used by SVDQuant.
At the same vector-storage budget as a BF16 rank-$32$ branch, \method{} uses rank $512$ binary factors and accounts for $47$--$53\%$ of the Frobenius norm of LLM weight matrices across three model sizes, compared with $26$--$29\%$ for BF16 rank $32$. 
For perplexity, \method{} matches or improves the BF16 rank-$32$ correction on  WikiText-2/C4 perplexity  across five models.
We implement \method{} on Blackwell GPUs as a \emph{fused single kernel} with the factors bit-packed at runtime: 
the native FP4 GEMM runs on tensor cores, the binary correction is computed on the CUDA cores of the same launch, and the correction is added inside the GEMM epilogue.
For batch-one inference, against the fastest FP4-only kernel we could build at the shape, \method{} adds $6$--$11\%$ latency for ranks $16$ to $512$. 
For larger batches, the cost of the binary branch on CUDA cores grows with the batch size, while the FP4 GEMM time, which is dominated by reading the weights, does not change for batch sizes up to $128$.
The largest batch size at which the overhead stays within $20\%$ therefore decreases with rank: $8$ at $r{=}128$, $4$ at $r{=}256$, and $2$ at $r{=}512$.
Our code is publicly available.\footnote{\url{https://github.com/jindavid/BiLoCo}}

%% file: sections/02_background.tex
\section{Background}
\label{sec:background}

\subsection{FP4 Quantization}
\label{sec:background_nvfp4}

Given the FP4 native tensor core GEMMs on the Blackwell generation of GPUs, there has been much recent work on investigating FP4 quantization. In this work we work with NVFP4 quantization, which partitions the input dimension of a tensor into blocks of $16$ elements~\cite{nvidiaIntroducingNVFP4}. 
Each block is quantized to E2M1 (4-bit, $1$ sign + $2$ exponent + $1$ mantissa) with a shared E4M3 (FP8) scale, and the whole tensor has one additional FP32 scale. 
For a weight matrix $W \in \mathbb{R}^{m \times n}$, the quantized representation is $\hat W_{ij} = g\,s_{i,\lfloor j/16 \rfloor}\,q_{ij}$ with $q_{ij} \in \{0,\pm 0.5,\pm 1,\pm 1.5,\pm 2,\pm 3,\pm 4,\pm 6\}$, $s$ the per-block FP8 scale, and $g$ the per-tensor FP32 scale. 
The on-disk cost is approximately $4.5$ bits per weight ($4$ weight bits plus $0.5$ scale bits). 

NVFP4 is attractive because both storage and arithmetic are low precision. When
the activation and weight operands are in the native block-scaled FP4 format,
Blackwell tensor cores can execute the main GEMM directly. However, the E2M1
codebook is extremely coarse, and each block has only one local scale. A few
large entries in a block can therefore increase error for the remaining entries,
while aggressive activation quantization can also amplify the effect of hidden
state outliers. Thus, direct PTQ to NVFP4 can still leave a measurable accuracy
gap, especially when both weights and activations are quantized.

\subsection{Low-Rank Corrections for Quantization Error}
\label{sec:background_svdquant}
One way to reduce this gap is to decompose a weight matrix into a low-bit
residual plus a higher-precision correction.
For weight-only quantization, LQ-LoRA \citep{guo2024lqloralowrankplusquantized} and LoftQ \citep{li2023loftq} decompose pretrained weights into a
quantized component and a high-precision low-rank component via SVD, then fine-tune the low-rank component.  

SVDQuant applies this decomposition for 4-bit weight-and-activation
quantization for diffusion models~\cite{li2025svdquant}. It first migrates
activation outliers into the weights, then uses SVD to decompose a matrix $W$ into a  low-rank component $VU^\top$ and a residual $R$, which is  quantized to 4 bits. Inference proceeds as 
\begin{equation}
    y
    =
    \text{FP4-GEMM}\!\left(
        \text{FP4}(x),\text{FP4}(R)
    \right)
    +
    x V U^\top.
    \label{eq:two_branch_background}
\end{equation}

This correction is effective because SVD captures directions of large residual
energy that are poorly represented by the 4-bit residual. It also has a clear
cost, since the correction stores BF16 left and right factors and is thus naturally
implemented as one or more higher-precision GEMMs. 

\subsection{Tensor Cores and CUDA Cores}
\label{sec:background_gpu}
Each streaming multiprocessor (SM) of an NVIDIA GPU has two kinds of arithmetic units: tensor cores, which execute matrix multiply-accumulate instructions, and general-purpose CUDA cores, which execute scalar and vector instructions such as additions and comparisons.
The two kinds of units issue instructions independently, so one group of threads can keep the tensor cores busy with a GEMM while another group uses the CUDA cores.
FlashAttention-3 and FlashAttention-4 use this to overlap the softmax, computed on CUDA cores and special function units, with the attention GEMMs on tensor cores~\cite{shah2024flashattention3fastaccurateattention,zadouri2026flashattention4algorithmkernelpipelining}.
A GPU kernel runs as a grid of thread blocks, or cooperative thread arrays (CTAs), each scheduled on one SM and divided into warps of $32$ threads.
In a CUTLASS GEMM on Blackwell, each CTA computes one tile of the output.
Some of its warps load the operands and issue tensor-core instructions in the main loop, and the remaining epilogue warps convert the accumulators and write the output tile to memory.
At LLM decode batch sizes the GEMM is limited by reading the weights from memory, so the tensor cores are underused and the epilogue warps wait during most of the main loop.

%% file: sections/03a_method.tex
\section{Method}
\label{sec:method}

\subsection{Motivating Study: LLM Weights Favor Rank Multiplicity}
\label{sec:llm_structure}
\label{sec:capture}

For a given memory budget, should a correction use a few high-precision rank-one components or many low-precision ones?
For an $m \times n$ matrix, a BF16 rank-$k$ correction stores $2k$ vectors in 16 bits, for $16k(m+n)$ bits.
A binary rank-$r$ correction stores $r$ sign vectors per side and $r$ BF16 scales, for $r(m+n+16)$ bits.
At equal storage, the binary correction can therefore use about $16$ times as many components.
More components help only if the spectrum of $W$ is flat enough that many directions matter, which we test next.
\begin{figure}[t]
\centering
\begin{minipage}[c]{0.5\textwidth}
\centering
\includegraphics[width=\linewidth]{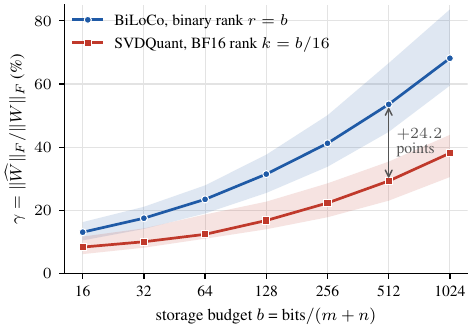}
\end{minipage}\hfill
\begin{minipage}[c]{0.46\textwidth}
\caption{Fraction $\gamma=\|\widehat W\|_F/\|W\|_F$ of the Frobenius norm in the low-rank part versus the storage budget $b$, on $63$ Qwen3-4B projection matrices ($7$ projection types in $9$ decoder layers).
At each budget, \method{} uses binary rank $r=b$ and SVDQuant uses BF16 rank $k=b/16$, so both use the same storage.
Lines are medians and bands are interquartile ranges.}
\label{fig:capture_vs_rank}
\end{minipage}
\end{figure}
We test this on $189$ projection matrices from Qwen3-4B, Qwen3-14B, and Llama-3-8B, including $9$ decoder layers per model and $7$ projection types per layer. 
For each matrix, we compare BF16 SVD rank $32$ with a signed rank-$512$. 
We measure $\gamma = \|\widehat W\|_F/\|W\|_F$, the fraction of the Frobenius norm of $W$ in the low-rank part $\widehat W$.
Table~\ref{tab:capture_summary} shows the results, where we find that the top
spectrum is flat, with the median $\sigma_1/\sigma_{32}$
ranging from $1.93$ on Qwen3-4B to $2.24$ on Llama-3-8B. At equal vector
storage, signed rank $512$ retains a median $47$--$53\%$ of
$\|W\|_F$, versus $26$--$29\%$ for BF16 SVD rank $32$. 
The ratio of the two values of $\gamma$ is anti-correlated with $\sigma_1/\sigma_{32}$ on every family, matching the rank-multiplicity prediction.

Figure \ref{fig:capture_vs_rank} plots $\gamma$ against vector-storage budget $b$. 
At each matched budget, SVD uses BF16 rank $k$=$b$/16, while binary factorization uses signed rank $r=b$. 
The binary factorization has a higher $\gamma$ at every tested budget.
The advantage is not isolated to one projection type as shown in Table~\ref{tab:capture_by_projection}. 
This motivates \method{}, where the same correction bytes can cover many more directions if the factors are one-bit signs. 
Since $\gamma$ is only a proxy for model quality, Section~\ref{sec:exp} evaluates whether this storage advantage preserves model quality.

\begin{table}[t]
\caption{Fraction $\gamma$ at equal storage across three model
families. Signed rank $512$ and BF16 SVD rank $32$ use the same
vector-storage budget.}
\label{tab:capture_summary}
\centering
\small
\begin{tabular}{lrrrrr}
\toprule
model & $\sigma_1/\sigma_{32}$ med. & $\gamma$, SVD $k{=}32$ & $\gamma$, sign $r{=}512$ & ratio & Pearson \\
\midrule
Qwen3-4B    & $1.93$ & $29.3\%$ & $52.9\%$ & $1.92\times$ & $-0.65$ \\
Qwen3-14B   & $2.00$ & $26.3\%$ & $47.3\%$ & $1.77\times$ & $-0.65$ \\
Llama-3-8B  & $2.24$ & $27.9\%$ & $50.6\%$ & $1.76\times$ & $-0.81$ \\
\bottomrule
\end{tabular}
\end{table}

\subsection{\method{}: Binary Low-Rank Corrections for Quantized Weights}
\label{sec:method_correction}

\method{} uses binary low rank components to  factorize the weight matrix and stores the residuals in NVFP4. Concretely, \method{} uses the same two-branch inference form as SVDQuant
(Eq.~\ref{eq:two_branch_background}), but changes the representation of the
correction branch. Instead of a BF16 factorization, it stores signed
rank-one terms and one scalar coefficient per term.
At PTQ time, for each weight matrix $W$, we choose a correction
$C=\sum_i \alpha_i u_i v_i^\top$ and store the residual
$R = W-C$ in NVFP4. NVFP4 uses 16-element input-channel blocks,
E2M1 4-bit values, and a shared FP8 scale, for approximately
$4.5$ bits per residual weight. We represent the correction as
\begin{equation}
W \;\approx\; \sum_{i=1}^{r} \alpha_i \, u_i v_i^\top \;+\; R, \qquad u_i \in \{\pm 1\}^m, \; v_i \in \{\pm 1\}^n, \; \alpha_i \in \mathbb{R}
\label{eq:binary_decomp}
\end{equation}
The deployed forward pass, with BF16 coefficient $\alpha$, is
\begin{equation}
y \;=\; \text{FP4-GEMM}(\text{FP4}(x), \text{FP4}(R)) \;+\; \sum_{i=1}^{r} \alpha_i \, u_i \, (v_i^\top x).
\label{eq:binary_forward}
\end{equation}
Because each $v_{i} \in \{\pm 1\}^n$, the inner product $v_i^\top x$ requires only additions and subtractions, i.e., $\sum_k v_{i,k} \, x_k$ where $v_{i,k} \in \{\pm 1\}$
selects between $+x_k$ and $-x_k$. This extends to the outer product with $u_i$. 
The low-rank term therefore needs no multiplications, and its factors take $16\times$ less memory than BF16 factors.

To find the binary low-rank corrections, we avoid an explicit SVD and incorporate a modified signed-cut decomposition algorithm~\cite{Frieze1999, riasanovsky2024efficient1bittensorapproximations} to build $\{u_i, v_i, \alpha_i\}_{i=1}^r$
greedily. We first initialize $R \leftarrow W$. Then for each rank-one component, we
draw a random sign vector $v$, then alternate
\begin{equation}
u \leftarrow \mathrm{sign}(R v), \qquad v \leftarrow \mathrm{sign}(R^\top u)
\end{equation}
for $T$ iterations. The optimal scalar for fixed binary $u, v$ is $\alpha = (u^\top R v) / (mn)$, which minimizes $\|R - \alpha u v^\top\|_F^2$ subject to $u, v$ being binary. 
We then deflate via $R \leftarrow R - \alpha u v^\top$ and proceed on to the next rank.
In practice, $T = 5$ is sufficient for the alternation to settle.
\footnote{The sign alternation and the joint refit follow \citet{riasanovsky2024efficient1bittensorapproximations}. Our contributions are using them to correct NVFP4 quantization error and the fused kernel of \S\ref{sec:method_kernel}.}

\begin{algorithm}[t]
\small
\caption{Greedy binary peeling with global coefficient refit}
\label{alg:peel}
\begin{algorithmic}[1]
\Require Weight matrix $W \in \mathbb{R}^{m \times n}$,
rank $r$, alternation steps $T$
\State $R \gets W$
\For{$i = 1,\ldots,r$}
  \State initialize $v_i \in \{\pm1\}^n$ at random
  \For{$t = 1,\ldots,T$}
    \State $u_i \gets \mathrm{sign}(R v_i)$
    \State $v_i \gets \mathrm{sign}(R^\top u_i)$
  \EndFor
  \State $\alpha_i \gets (u_i^\top R v_i)/(mn)$
  \State $R \gets R - \alpha_i u_i v_i^\top$
\EndFor
\State Form $A_{ij}=(u_i^\top u_j)(v_i^\top v_j)$ and
$b_i=u_i^\top W v_i$
\State Solve $(A+\epsilon I)\alpha=b$ and set
$R \gets \text{FP4}(W-\sum_i \alpha_i u_i v_i^\top)$
\State \Return sign vectors $\{u_i,v_i\}_{i=1}^r$, BF16 coefficients
$\alpha$, and FP4 residual $R$
\end{algorithmic}
\end{algorithm}

Greedy peeling produces $(u_i, v_i)$ at a coordinate-wise stationary
point per term but uses the per-term closed form
$\alpha_i = (u_i^\top R_i v_i)/(mn)$ that ignores cross-term
correlations. Once all $r$ signed terms are produced, we refit
$\boldsymbol{\alpha} = \{\alpha_i\}_{i=1}^r$ jointly in closed form:
\begin{equation}
\boldsymbol{\alpha}^* = \arg\min_{\boldsymbol{\alpha} \in \mathbb{R}^r}
\Bigl\|W - \textstyle\sum_i \alpha_i u_i v_i^\top\Bigr\|_F^2,
\end{equation}
which solves the $r \times r$ system $A \boldsymbol{\alpha} = b$ with
$A_{ij} = (u_i^\top u_j)(v_i^\top v_j)$ and $b_i = u_i^\top W v_i$.
A small regularization $\epsilon I$ (i.e., as in ridge regression) is added for numerical stability. 
The solve is $O(r^3)$ and runs in milliseconds per layer. 
Building $b$ shares the $U^\top W V$ matmul that the peel step already amortizes.
After the refit, the storage residual $R_\mathrm{store}=W-U\mathrm{diag}(\alpha)V^\top$ is quantized to NVFP4 once. 
Algorithm~\ref{alg:peel} summarizes the full peeling and refit procedure. 

%% file: sections/03b_theory.tex
\subsection{Theoretical Analysis}
\label{sec:why_works}

\label{sec:algorithmic_analysis}

\method{} uses greedy sign-alternation peeling
because each coordinate update maximizes the current bilinear objective
over one set of signs, and therefore monotonically improves the
single-term Frobenius decrease. After the signs are fixed, the final
coefficient refit is the least-squares optimum over the selected rank-one components.
Our theoretical analysis shows that the PTQ solver is a principled matching-pursuit procedure for the Frobenius norm as a proxy objective. (The proofs are in Appendix~\ref{app:proofs}.)

\begin{theorem}[Monotone single-term sign alternation]
\label{thm:per_step}
Let $R \in \mathbb{R}^{m \times n}$ be fixed. Starting from any
$v^{(0)}\in\{\pm1\}^n$, alternate
$u^{(t)}=\mathrm{sign}(Rv^{(t-1)})$ and
$v^{(t)}=\mathrm{sign}(R^\top u^{(t)})$, with deterministic
tie-breaking. Then $u^\top Rv$ is non-decreasing after each
coordinate update. For the final signs $u, v$, the optimal scalar is
$\alpha^\star=(u^\top Rv)/(mn)$, and deflation yields
\[
  \|R\|_F^2 - \|R - \alpha^\star u v^\top\|_F^2
  \;=\; \frac{(u^\top R v)^2}{mn}.
\]
\end{theorem}

\begin{proposition}[Global coefficient refit]
\label{thm:alpha_refit}
For fixed sign matrices
$U \in \{\pm 1\}^{m \times r}$ and
$V \in \{\pm 1\}^{n \times r}$, the least-squares coefficients
\[
    \alpha^\star
    =
    \arg\min_\alpha
    \|W - \sum_i \alpha_i u_i v_i^\top\|_F^2
\]
solve $A\alpha=b$, where $A_{ij}=(u_i^\top u_j)(v_i^\top v_j)$ and $b_i=u_i^\top W v_i$.
Replacing the greedy coefficients by $\alpha^\star$ cannot increase
the Frobenius residual.
\end{proposition}

\paragraph{Cut-norm motivation.}
The rank-one components can also be viewed through the cut norm.
Let $\|R\|_\blacksquare=\max_{u\in\{\pm1\}^m,\,v\in\{\pm1\}^n} u^\top R v$ denote the signed cut norm of $R$.
If, at each peeling
step $t$, the selected signs satisfy
\[
    |u_t^\top R_t v_t|
    \geq
    \rho\,\|R_t\|_\blacksquare
\]
for some $\rho \in (0,1]$, then with
$\alpha_t=(u_t^\top R_t v_t)/(mn)$ and
$R_{t+1}=R_t-\alpha_tu_tv_t^\top$, the residuals satisfy
\[
    \min_{0 \leq t < r}
    \|R_t\|_\blacksquare^2
    \;\leq\;
    \frac{mn}{\rho^2 r}\,\|W\|_F^2 .
\]
This is a conditional Frieze--Kannan-style statement~\cite{Frieze1999, 10.1145/1007352.1007371}: it explains
what a uniformly good signed-cut oracle would imply, but it is not a claim that our alternating heuristic attains a fixed approximation ratio to the optimal cut norm. 
Empirically, we measure the rate per step
\[
    \rho_F(t)
    =
    \frac{|u_t^\top R_t v_t|}
    {\sqrt{mn}\,\|R_t\|_F},
\]
which is exactly the rate constant in
\[
    \|R_{t+1}\|_F^2
    =
    \|R_t\|_F^2
    \left(1-\rho_F(t)^2\right).
\]
Across $63$ Qwen3-4B matrices and $8064$ measured rank-one updates,
$\rho_F$ remains bounded away from zero in the diagnostic sweep. The full \(\rho_F\) distribution is reported in Appendix~\ref{app:structural_diagnosis}.

\paragraph{Scope of Analysis.}
The analysis establishes three local facts: the per-term sign search is monotone, the deployed coefficients are optimal over the chosen
signs, and a conditional cut-norm oracle would yield a standard residual-decay statement.
It does not bound output error, downstream
accuracy, or PPL.
The role of the analysis is to support the storage--accuracy argument from Section~\ref{sec:llm_structure} and to make the greedy signed-peeling solver principled rather than ad hoc. The model-quality claims are validated empirically in Section~\ref{sec:exp}.

%% file: sections/03c_kernel.tex
\subsection{Kernel Design}
\label{sec:method_kernel}
We realize the efficiency argument of \method{} with a \emph{fused
single kernel}, which runs the FP4 GEMM on tensor cores while computing the bit-packed binary correction on CUDA cores.
For a layer with $W\in\mathbb{R}^{N\times K}$,
input $X\in\mathbb{R}^{M\times K}$, and output
$Y\in\mathbb{R}^{M\times N}$, we have
\begin{equation}
    Y =
    \underbrace{X\,R^\top}_{\text{tensor-core path}}
    +
    \underbrace{\sum_{i=1}^{r}\alpha_i (Xv_i)u_i^\top}_{\text{\method{} branch}},
    \qquad
    u_i\in\{\pm1\}^N,\; v_i\in\{\pm1\}^K .
    \label{eq:branch_split}
\end{equation}

\paragraph{Fused execution.}
The kernel is built on a CUTLASS SM100 block-scaled FP4 GEMM and launches at most one thread block (cooperative thread array, CTA) per streaming multiprocessor (SM).
The GEMM uses output tiles of $128\times64$, with the batch zero-padded to $128$ rows, so the first $N/64$ CTAs each compute one output tile with the unmodified tensor-core GEMM.
Because these CTAs wait for the correction inside the kernel, all CTAs must run at the same time, which requires $N/64$ to be smaller than the number of SMs ($148$ on B300).
The remaining CTAs never touch the GEMM.
They compute the branch on CUDA cores directly from the packed sign bits, where each inner product $v_i^\top x$ is a sequence of additions whose signs come from the bits of $v_i$. 
Each scaled result $z_i=\alpha_i(v_i^\top x)$ is written to a buffer in global memory together with the index of the current launch (an epoch tag).
The buffer is reused across launches, and the tag lets a reader tell a value written by the current launch from a stale one, so the buffer never has to be cleared, which would cost an extra kernel launch.
Every output tile needs all $r$ values of $z$.
In each GEMM CTA, the epilogue warps, which are idle during the main loop, therefore wait until $z$ has been written, expand it through the packed $u_i$ into the tile's correction, and add it to the FP32 accumulators before the BF16 stores.
The low-rank product is thus split in two: the side CTAs compute $z$ on SMs the GEMM does not use, and the epilogue warps compute each tile's part of $\sum_i z_i u_i^\top$ while the tensor cores run the main loop.
By scheduling the correction onto idle SMs and executing the addition inside the GEMM epilogue, BiLoCo avoids separate kernel launches, CUDA streams, or auxiliary memory passes.
Implementation details are in Appendix~\ref{app:kernel_details}.

\paragraph{Memory and precision.}
Both sign factors stay bit-packed at runtime, with $V\in\{0,1\}^{r\times K/8}$ and $U\in\{0,1\}^{r\times N/8}$. 
At $r{=}128$ the branch reads about $0.16$\,MiB of signs plus the BF16 token, while the FP4 path reads ${\sim}12.5$\,MiB of E2M1 weights and $1.6$\,MiB of scales, 
so the correction moves $80{\times}$ less data.
Its cost is the time of a chain of dependent steps (summing over $K$, publishing $z$, waiting for it in the GEMM CTA, and expanding it), not the data it reads.

\paragraph{Batch size and rank.}
The same single kernel runs at every batch size and rank we measure.
The work of the branch grows with $M r$, while the FP4 GEMM time, dominated by reading the weights, does not change for $M\le128$.
The branch is hidden as long as it finishes during the GEMM main loop, so the batch range over which it stays hidden depends on the rank (\S\ref{sec:exp_kernel}).
For $M{=}16$ with $r{\geq}256$ and $M{=}32$ with $r{\geq}128$, the side CTAs alone are too slow, so every CTA first computes the correction and the GEMM runs afterwards.
This is still one launch, but the correction is no longer overlapped with the GEMM.
Beyond that batch range, and at the wide MLP projections where $N/64$ exceeds the SM count, a tensor-core correction is the better choice.

%% file: sections/05_experiments.tex
\section{Experiments}
\label{sec:exp}

We evaluate \method{} in two parts.
\S\ref{sec:exp_accuracy} asks whether \method{} improves model quality, measured by perplexity, KL divergence to the BF16 model, and downstream task performance over FP4 post-training quantization.
\S\ref{sec:exp_kernel} then measures the decode latency of the fused kernel of \S\ref{sec:method_kernel} against the FP4-only kernel across batch sizes.

\subsection{Performance Evaluation}
\label{sec:exp_accuracy}

\paragraph{Setup.}
We evaluate on Qwen3-4B, Llama-3-8B, Qwen3-14B,  Qwen3-32B, and Qwen2.5-72B. All quantization uses SmoothQuant \citep{xiao2024smoothquantaccurateefficientposttraining} presmoothing with $\alpha = 0.5$, $T = 5$ sign-alternation steps per rank-one component, and the ``four over six'' NVFP4 per-block MSE scale search following \citet{cook2025sixaccuratenvfp4quantization} for both weights and activations. For \(r{=}512\), the full \method{} PTQ pass takes approximately 1 minute for Qwen3-4B and 5 minutes for Qwen3-32B on a single B300.
Calibration uses $128$ sequences of length $2048$ from the C4 training set.
SVDQuant is deterministic given this calibration set.
The WikiText-2 test set is concatenated and split into non-overlapping $2048$-token windows, and C4 validation uses $256$ fixed-seed random $2048$-token windows.

We compare \method{} and SVDQuant at the same storage budget $b{=}512$ on perplexity (Table~\ref{tab:ppl_b512}), KL divergence to the BF16 model (Table~\ref{tab:kl}), and downstream accuracy (Table~\ref{tab:downstream_openllm}).
We also report \method{} at $r{=}128$, which uses a quarter of that storage and has a $\gamma$ at least as large as SVDQuant $k{=}32$ (Table~\ref{tab:capture_by_projection}).
The appendix also reports perplexity across ranks (Appendix~\ref{app:solver_ablation}), per-task downstream accuracy (Appendix~\ref{app:full_downstream}), and KL divergence at $b{=}128$ (Appendix~\ref{app:full_kl}).

\begin{table}[t]
\caption{Perplexity of BiLoCo vs. baseline methods. BiLoCo with $b{=}512$ is memory-matched to SVDQuant writh $k=32$.
\method{} entries report mean$\,\pm\,$std over sign initializations.
Bold marks the best corrected FP4 method, with ties bolded.}
\label{tab:ppl_b512}
\centering
\small
\begin{tabular}{llrrrrr}
\toprule
Model & Dataset & BF16 & NVFP4
  & \makecell{SVDQuant\\$k{=}32$}
  & \makecell{\method{}\\$r{=}128$}
  & \makecell{\method{}\\$r{=}512$} \\
\midrule
Qwen3-4B   & Wiki2 & $14.22$ & $15.13$ & $15.07$ & $15.09{\scriptstyle\,\pm 0.58}$ & $\mathbf{14.94}{\scriptstyle\,\pm 0.37}$ \\
Qwen3-4B   & C4    & $15.25$ & $15.88$ & $15.85$ & $15.88{\scriptstyle\,\pm 0.23}$ & $\mathbf{15.81}{\scriptstyle\,\pm 0.17}$ \\
\midrule
Llama-3-8B & Wiki2 &  $6.27$& $7.00$ & $6.92$ & $6.93{\scriptstyle\,\pm 0.01}$ & $\mathbf{6.84}{\scriptstyle\,\pm 0.01}$ \\
Llama-3-8B & C4    &  $6.69$ & $7.68$ & $7.59$ & $7.61{\scriptstyle\,\pm 0.01}$ & $\mathbf{7.48}{\scriptstyle\,\pm 0.01}$ \\
\midrule
Qwen3-14B  & Wiki2 & $8.89$ & $9.36$ & $9.19$ & $9.21{\scriptstyle\,\pm 0.03}$ & $\mathbf{9.17}{\scriptstyle\,\pm 0.04}$ \\
Qwen3-14B  & C4    & $11.19$ & $11.55$ & $11.46$ & $11.49{\scriptstyle\,\pm 0.03}$ & $\mathbf{11.44}{\scriptstyle\,\pm 0.02}$ \\
\midrule
Qwen3-32B  & Wiki2 & $7.85$ & $8.23$ & $8.18$ &  $8.23{\scriptstyle\,\pm 0.03}$ & $\mathbf{8.16}{\scriptstyle\,\pm 0.03}$ \\
Qwen3-32B  & C4    & $10.09$  & $10.46$ & $\mathbf{10.40}$ & $10.44{\scriptstyle\,\pm 0.01}$ & $\mathbf{10.40}{\scriptstyle\,\pm 0.02}$ \\
\midrule
Qwen2.5-72B  & Wiki2 & $3.92$ & $4.29$ & $\mathbf{4.23}$ &  $4.26{\scriptstyle\,\pm 0.01}$ & $\mathbf{4.23}{\scriptstyle\,\pm 0.01}$ \\
Qwen2.5-72B  & C4    & $7.83$  & $8.10$ & $8.05$ & $8.07{\scriptstyle\,\pm 0.01}$ & $\mathbf{8.00}{\scriptstyle\,\pm 0.01}$ \\
\bottomrule
\end{tabular}
\vspace{-3mm}
\end{table}

\paragraph{Perplexity.}
\method{} entries are means over results from six sign initializations, and the table reports their standard deviation. 
At $b{=}512$, \method{} ($r$=512) has lower perplexity than SVDQuant ($k$=32) on both datasets for Qwen3-4B/14B and Llama-3-8B.
On Qwen3-32B, \method{} is slightly better on WikiText-2 and tied on C4.
On Qwen2.5-72B, it is tied on WikiText-2 and better on C4.
At $r{=}128$, which uses a quarter of the storage, \method{} is slightly behind SVDQuant, by at most $0.05$ PPL, on every model and dataset.

\paragraph{KL divergence.}
We compare the quantized next-token distributions to the BF16 reference model on $200$ held-out FineWeb documents of up to $2048$ tokens (Table~\ref{tab:kl}).
We report the mean KL over all tokens and the 99th percentile of the per-document mean KL.
At $b{=}512$, \method{} $r{=}512$ has lower 99th-percentile KL than SVDQuant on all five models and lower mean on all but Qwen2.5-72B.
On four of the five models the relative gap is larger at the 99th percentile (on Qwen3-14B, $0.130$ vs.\ $0.167$, $22\%$ lower), so the improvement is not limited to typical documents.
Appendix~\ref{app:full_kl} also compares the two methods at $b{=}128$ (\method{} $r{=}128$ vs.\ SVDQuant $k{=}8$).

\begin{table}[t]
\vspace{-2mm}
\caption{KL divergence to the BF16 model at $b{=}512$ on FineWeb document. Lower is better.}
\label{tab:kl}
\centering
\small
\begin{tabular}{lcccc}
\toprule
& \multicolumn{2}{c}{KL mean ($\downarrow$)}
& \multicolumn{2}{c}{KL p99 ($\downarrow$)} \\
\cmidrule(lr){2-3}\cmidrule(lr){4-5}
Model & \method{} $r{=}512$ & SVDQuant $k{=}32$ & \method{} $r{=}512$ & SVDQuant $k{=}32$ \\
\midrule
Qwen3-4B    & $\mathbf{0.0740}$ & $0.0835$ & $\mathbf{0.1421}$ & $0.1649$ \\
Qwen3-14B   & $\mathbf{0.0480}$ & $0.0517$ & $\mathbf{0.1303}$ & $0.1671$ \\
Llama-3-8B  & $\mathbf{0.0756}$ & $0.0824$ & $\mathbf{0.3252}$ & $0.3409$ \\
Qwen3-32B   & $\mathbf{0.0589}$ & $0.0605$ & $\mathbf{0.3175}$ & $0.3363$ \\
Qwen2.5-72B   & $0.0515$ & $\mathbf{0.0497}$ & $\mathbf{0.1937}$ & $0.2097$ \\
\bottomrule
\end{tabular}
\vspace{-2mm}
\end{table}

\paragraph{Downstream tasks.}
Table~\ref{tab:downstream_openllm} reports the WinoGrande (5-shot) and HellaSwag (10-shot) mean accuracy, following the protocol of \citet{egiazarian2026bridginggappromiseperformance}. 
At $b{=}512$, \method{} improves over SVDQuant on the four Qwen models by $0.1$--$0.7$ points and is $0.3$ lower on Llama-3-8B. 
With $r{=}128$, which uses a quarter of the storage, \method{} is $0.4$ points above SVDQuant $k{=}32$ on Qwen3-14B and at most $0.3$ points below it on every model.
Appendix~\ref{app:full_downstream} reports per-task results on five standard PTQ tasks (ARC-easy, ARC-challenge, PIQA, WinoGrande, OBQA).
There, \method{} $r{=}512$ has a higher mean than SVDQuant on Qwen3-4B and Qwen3-14B and is $0.3$ points lower on Llama-3-8B, where \method{} $r{=}128$ has the highest mean.

\begin{table}[t]
\caption{Mean accuracy on WinoGrande (5-shot) and HellaSwag (10-shot). Higher is better.}
\label{tab:downstream_openllm}
\centering
\small
\begin{tabular}{lccccc}
\toprule
Method                        & Qwen3-4B & Qwen3-14B & Llama-3-8B & Qwen3-32B & Qwen2.5-72B \\
\midrule
BF16 (reference)              & $0.686$ & $0.772$ & $0.803$ & $0.803$ & $ 0.857$\\
\midrule
SVDQuant $k{=}32$   ($b{=}512$)          & $0.669$ & $0.759$ & $\mathbf{0.780}$ & $0.796$ &  $ 0.849$ \\
\method{} $r{=}128$ ($b{=}128$) & $0.667$ & $\mathbf{0.763}$ & $0.777$ & $0.795$ & $ 0.846$ \\
\method{} $r{=}512$ ($b{=}512$) & $\mathbf{0.670}$ & $\mathbf{0.763}$ & $0.777$ & $\mathbf{0.799}$ & $ \mathbf{0.856}$\\
\bottomrule
\end{tabular}
\vspace{-3mm}
\end{table}

\subsection{Decode Kernel Efficiency}
\label{sec:exp_kernel}
\paragraph{Setup.}
Kernel measurements run on NVIDIA B300 (SM103, 148 SMs) at the decode shape $K{=}N{=}5120$ for batch sizes $M{=}1$--$32$. 
We capture $64$ launches in one CUDA graph and report the median time per launch over $15$ interleaved passes. 
Timing one launch at a time would add launch overhead comparable to the kernel itself.
Every ratio is computed against the fastest FP4-only GEMM kernel with the same $128{\times}64$ tiles and the activation zero-padded to $128$ rows, measured in the same run, because
FP4 GEMM time is data-content sensitive at small $M$ and cross-run
ratios are not meaningful. 
The padded kernel is the fastest FP4-only kernel we found at this shape ($5.8$--$5.9\,\mu$s).
The fastest unpadded FP4-only GEMM is $23$--$36\%$ slower, so ratios against it would make the \method{} kernel look faster than it is.

\begin{table}[t]
\centering
\small
\providecommand{\fast}[2]{\makecell{$#1$\\[-1pt]{\scriptsize $#2\times$}}}
\providecommand{\slow}[2]{\makecell{\color{gray}$#1$\\[-1pt]\color{gray}{\scriptsize $#2\times$}}}
\caption{Decode latency of the \method{} kernel on B300 ($K{=}N{=}5120$). Each entry gives the time per launch in $\mu$s and, below it, the ratio to the padded FP4-only kernel measured in the same run ($5.8$--$5.9\,\mu$s). Gray entries are more than $20\%$ slower than the FP4-only kernel.}
\label{tab:binarysvd_kernel}
\renewcommand{\arraystretch}{1.1}
\setlength{\tabcolsep}{11pt}
\begin{tabular*}{\textwidth}{@{\extracolsep{\fill}}c cccccc@{}}
\toprule
& \multicolumn{6}{c}{Rank $r$} \\
\cmidrule(l){2-7}
Batch $M$ & $16$ & $32$ & $64$ & $128$ & $256$ & $512$ \\
\midrule
  $1$  & \fast{6.17}{1.06} & \fast{6.18}{1.06} & \fast{6.23}{1.06} & \fast{6.33}{1.08} & \fast{6.39}{1.09} & \fast{6.49}{1.11} \\[2pt]
  $2$  & \fast{6.36}{1.08} & \fast{6.36}{1.09} & \fast{6.36}{1.09} & \fast{6.32}{1.08} & \fast{6.36}{1.09} & \fast{6.71}{1.15} \\[2pt]
  $4$  & \fast{6.36}{1.09} & \fast{6.36}{1.09} & \fast{6.46}{1.10} & \fast{6.47}{1.11} & \fast{6.90}{1.18} & \slow{8.45}{1.44} \\[2pt]
  $8$  & \fast{6.42}{1.10} & \fast{6.42}{1.10} & \fast{6.82}{1.17} & \fast{6.74}{1.15} & \slow{8.02}{1.37} & \slow{10.10}{1.74} \\[2pt]
  $16$ & \fast{6.55}{1.11} & \fast{6.52}{1.11} & \slow{7.48}{1.28} & \slow{9.08}{1.55} & \slow{14.70}{2.51} & \slow{17.81}{3.04} \\[2pt]
  $32$ & \fast{6.90}{1.18} & \slow{8.86}{1.51} & \slow{11.46}{1.96} & \slow{15.03}{2.56} & \slow{16.94}{2.90} & \slow{19.33}{3.30} \\
\bottomrule
\end{tabular*}
\end{table}
\vspace{-2mm}

\paragraph{Fused kernel per batch size/rank.}
Table~\ref{tab:binarysvd_kernel} reports decode latency across batch sizes and ranks.
At batch size $M{=}1$, the correction adds $6$--$11\%$ over the FP4-only kernel for every rank from $16$ to $512$ ($0.34$--$0.64\,\mu$s) in single launch.
Because the binary branch runs on CUDA cores, its cost grows with $M$ and $r$ while the FP4 GEMM stays flat, so the batch range over which \method{} stays within $20\%$ of FP4-only declines with larger ranks (from $M{=}32$ at $r{=}16$ to $M{=}2$ at $r{=}512$ in Table~\ref{tab:binarysvd_kernel}).
Beyond those points the binary branch is no longer hidden, and we expect a tensor-core correction to be the better choice.
At equal rank, a dense BF16 correction placed in the same kernel costs the same as the binary one (Appendix~\ref{app:equal_rank}). 
The advantage of the binary format is the higher rank that a fixed byte budget buys.

\paragraph{Fused kernel per layer shape.}
Table~\ref{tab:shape_fused} reports latency at $M{=}1$ for the three supported GEMM shapes in a Qwen3-14B decoder layer, at $r{=}128$, $256$, and $512$, against the faster of the padded and unpadded FP4-only GEMM of the same shape measured in the same run.
The overhead is smallest at the down projection, where the FP4 GEMM itself is longest, and largest at the narrow k/v projection.
The kernel does not support the gate and up projections, because their $N/64{=}272$ output tiles exceed the $148$ SMs.

\begin{table}[t]
\centering
\small
\providecommand{\fast}[2]{\makecell{$#1$\\[-1pt]{\scriptsize $#2\times$}}}
\providecommand{\slow}[2]{\makecell{\color{gray}$#1$\\[-1pt]\color{gray}{\scriptsize $#2\times$}}}
\caption{\method{} kernel across Qwen3-14B projection shapes ($M{=}1$, B300). Each entry gives the time per launch in $\mu$s and, below it, the ratio to the fastest FP4-only GEMM of the same shape measured in the same run.}
\label{tab:shape_fused}
\renewcommand{\arraystretch}{1.0}
\begin{tabular*}{\textwidth}{@{\extracolsep{\fill}}llcc ccc@{}}
\toprule
Shape ($K{\times}N$) & Role & padded & unpadded & $r{=}128$ & $r{=}256$ & $r{=}512$ \\
\midrule
$5120{\times}5120$  & attn q/o & $5.85$ & $7.54$ & \fast{6.33}{1.08} & \fast{6.39}{1.09} & \fast{6.49}{1.11} \\[2pt]
$5120{\times}1024$  & attn k/v & $5.72$  & $5.62$ & \fast{6.10}{1.09} & \fast{6.55}{1.17} & \fast{7.00}{1.25} \\[2pt]
$17408{\times}5120$ & MLP down & $12.42$ & $22.65$ & \fast{12.93}{1.04} & \fast{12.97}{1.04} & \fast{13.18}{1.06} \\
\bottomrule
\end{tabular*}
\end{table}

%% file: sections/06_related_work.tex
\section{Related Work}
\label{sec:related}

\paragraph{Post-training quantization for LLMs.}
GPTQ~\cite{frantar2023gptqaccurateposttrainingquantization} and AWQ~\cite{lin2026awqactivationawareweightquantization} target W4A16 (4-bit weights, BF16 activations).
SmoothQuant~\cite{xiao2024smoothquantaccurateefficientposttraining} introduced the per-channel scale migration we use as a preprocessing step.
QuaRot~\cite{ashkboos2024quarotoutlierfree4bitinference} and SpinQuant~\cite{liu2025spinquantllmquantizationlearned} are rotation-based W4A4 methods that pre-rotate weights and activations to make them more quantization-friendly. 
Rotation is orthogonal to correction format: our binary branch can be applied on top of a rotated backbone, and Appendix~\ref{app:rotation} evaluates it with QuaRot and SpinQuant on Llama-3-8B.

\paragraph{Low-rank corrections for FP4 quantization.}
SVDQuant~\cite{li2025svdquant} introduced the FP4-with-low-rank-correction recipe for diffusion models, with a BF16 rank-$32$ projector by default. 
We extend this pattern to LLMs and study an alternative format under the same byte budget, fitting $16 \times 32 = 512$ ${\pm}1$ binary rank-one components in place of a BF16 rank-$32$ correction. 
As demonstrated in Section~\ref{sec:capture}, LLM weight matrices exhibit flat top spectra, meaning that allocating a fixed memory budget toward higher rank coverage yields greater fidelity than allocating it toward the precision of each rank-one component.
LQ-LoRA~\cite{guo2024lqloralowrankplusquantized} explored low-rank + low-bit decomposition for fine-tuning, focused on training-time storage rather than inference latency. 

\paragraph{1-bit and ternary networks.}
BitNet~\cite{wang2023bitnetscaling1bittransformers} and 1.58-bit LLMs~\cite{zhang2026sparsebitnet158bitllmsnaturally} train models entirely in binary or ternary weights from scratch. They are not post-training corrections to a higher-precision model. 
We borrow the binary representation but apply it as an additive PTQ branch on top of FP4.

\paragraph{Signed cut decompositions on LLM weights.}
The closest  line of work~\cite{riasanovsky2024efficient1bittensorapproximations} applies Frieze--Kannan-style signed cut decompositions to LLM weight matrices. 
Their Algorithm 5 (greedy sign-alternation peel) is the same per-component search used inside our Algorithm~\ref{alg:peel}, and their Algorithm 6 is the global $\alpha$ refit. 
We use binary rank-one components as a low-rank \emph{additive} correction to an FP4 main path, framed by the matched-storage argument of Section~\ref{sec:capture}, and run the correction on CUDA cores inside the FP4 GEMM kernel.

\paragraph{Spectral structure of LLM weights.}

Prior work has used singular-value spectra to analyze or compress transformer weights: random-matrix analyses study where information is stored across the spectrum, and SVD-based compression methods quantify the loss from truncating singular components~\cite{staats2025smallsingularvaluesmatter,wang2025svdllmtruncationawaresingularvalue,wang2025svdllmv2optimizingsingular,yuan2025asvdactivationawaresingularvalue}.
Our Section~\ref{sec:capture} measurement is complementary. Rather than using SVD for compression directly, we measure the top-spectrum shape at the SVDQuant rank and find that it is flat enough that a fixed correction budget is better spent on many binary rank-one components than on fewer high-precision ones.

\paragraph{Heterogeneous execution on GPUs.}
Recent GPU kernels increasingly overlap matrix multiplication, memory movement, and other work in attention and serving systems~\cite{shah2024flashattention3fastaccurateattention,zadouri2026flashattention4algorithmkernelpipelining,lin2025qservewakv,9917312}. \method{} applies this heterogeneous design pattern within a single projection layer by assigning the residual GEMM to FP4 tensor cores and the binary correction to CUDA cores.

%% file: sections/07_limitations.tex
\section{Limitations}
\label{sec:limitations}

\paragraph{Kernel scope.}
The fused single kernel covers every batch size and rank we measured. 
It stays within $20\%$ of the FP4-only kernel through $M{=}8$ for ranks up to $r{=}128$, through $M{=}4$ at $r{=}256$, and through $M{=}2$ at $r{=}512$ (Table~\ref{tab:binarysvd_kernel}). 
Larger batches, and the wide MLP projections whose tile count exceeds the SM count, are better served by a tensor-core correction.
The KV cache stays BF16 throughout our experiments, as \method{} specifically targets linear weight projections and their activations rather than attention operations or KV-cache memory traffic. 
Our latency claims are scoped to B300 and the projection shapes of Tables~\ref{tab:binarysvd_kernel} and~\ref{tab:shape_fused}.

\paragraph{Composition with rotations.}
Our rotation experiments (Appendix~\ref{app:rotation}) are limited to a single model and seed, where both tested rotation techniques underperform presmoothing under NVFP4 block quantization.

%% file: sections/08_conclusion.tex
\section{Conclusion}

We presented \method{}, a post-training quantization method that splits each LLM weight matrix into a low-rank component with one-bit sign factors and an FP4 residual.
At a fixed storage budget,  many one-bit rank-one components describe LLM weights better than a few high-precision ones, and \method{} surpasses or matches the performance of SVDQuant at the same storage across models from 4B to 72B parameters.
The binary format simplifies the arithmetic of the low-rank branch into additions and subtractions, allowing it to execute on the idle CUDA cores concurrently with the tensor core GEMM.
We turned this into a single fused kernel on Blackwell, in which idle SMs compute the low-rank product from the packed signs and the GEMM epilogue adds it to the output.
Because the fused kernel closely matches baseline FP4 latency at small decode batch sizes, BiLoCo recovers significant quantization accuracy with minimal runtime overhead.

\section*{Acknowledgments}
This study was supported by a grant from Korea Technology and Information Promotion Agency (TIPA) for SMEs  (with LLOYKD) and MIT-IBM Computing Research Lab.
David Jin is supported by an Amazon PhD Fellowship.
We thank Modal for a compute grant that supported the experiments in this work.

%% file: sections/a2_additional_experiments.tex
\section{Additional Experimental Details}
\label{app:additional_experiments}

\subsection{Structural and cut-premise diagnostics}
\label{app:structural_diagnosis}
Aggregating the per-matrix capture diagnostic of
Section~\ref{sec:capture} by projection type shows that the binary advantage holds for every projection type on all three models. 
Table~\ref{tab:capture_by_projection} reports medians across $9$ evenly-spaced decoder layers per model.
The metric is the Frobenius capture $\mathrm{capA}(r) = \|\widehat{W}_r\|_F / \|W\|_F$ from Table~\ref{tab:capture_summary}, where $\widehat{W}_r$ is the rank-$r$ truncated SVD reconstruction
for SVDQuant and the rank-$r$ greedy signed-peel
reconstruction for \method{}.

\method{} $r{=}256$ captures more than SVDQuant $k{=}32$ on all $21$ projection--model pairs while using half the storage.
At $b{=}128$, a quarter of the storage, \method{} captures about the same fraction as SVDQuant on Qwen3-14B and Llama-3-8B and substantially more on Qwen3-4B.
This pattern follows the differing spectral flatness of the three families (Table~\ref{tab:capture_summary}).

\begin{table}[t]
\caption{Frobenius-norm capture $\mathrm{capA}(r) = \|\widehat{W}_r\|_F / \|W\|_F$ per projection type, as the median over $9$ evenly spaced decoder layers per model. SVDQuant uses rank $k{=}32$ ($b{=}512$), and \method{} uses $r{=}128$ ($b{=}128$, a quarter of that storage) and $r{=}256$ ($b{=}256$, half). Bold marks \method{} $r{=}256$, which is the highest in every row. Higher is better.}
\label{tab:capture_by_projection}
\centering
\small
\begin{tabular}{llrrr}
\toprule
Model & Projection & SVDQuant $k{=}32$ & \method{} $r{=}128$ & \method{} $r{=}256$ \\
\midrule
\multirow{7}{*}{Qwen3-4B}
 & q\_proj    & $28.1\%$ & $35.4\%$ & $\mathbf{44.6\%}$ \\
 & k\_proj    & $36.9\%$ & $43.5\%$ & $\mathbf{55.8\%}$ \\
 & v\_proj    & $24.7\%$ & $38.3\%$ & $\mathbf{51.8\%}$ \\
 & o\_proj    & $21.5\%$ & $30.8\%$ & $\mathbf{41.7\%}$ \\
 & gate\_proj & $18.5\%$ & $26.5\%$ & $\mathbf{34.9\%}$ \\
 & up\_proj   & $15.0\%$ & $23.4\%$ & $\mathbf{32.3\%}$ \\
 & down\_proj & $17.4\%$ & $24.1\%$ & $\mathbf{33.0\%}$ \\
\midrule
\multirow{7}{*}{Qwen3-14B}
 & q\_proj    & $32.1\%$ & $31.0\%$ & $\mathbf{39.5\%}$ \\
 & k\_proj    & $39.7\%$ & $40.9\%$ & $\mathbf{53.1\%}$ \\
 & v\_proj    & $29.0\%$ & $35.1\%$ & $\mathbf{47.9\%}$ \\
 & o\_proj    & $25.0\%$ & $25.5\%$ & $\mathbf{34.8\%}$ \\
 & gate\_proj & $22.6\%$ & $22.0\%$ & $\mathbf{28.6\%}$ \\
 & up\_proj   & $16.5\%$ & $17.4\%$ & $\mathbf{24.0\%}$ \\
 & down\_proj & $18.1\%$ & $19.3\%$ & $\mathbf{26.0\%}$ \\
\midrule
\multirow{7}{*}{Llama-3-8B}
 & q\_proj    & $35.1\%$ & $34.4\%$ & $\mathbf{44.0\%}$ \\
 & k\_proj    & $41.3\%$ & $43.2\%$ & $\mathbf{55.5\%}$ \\
 & v\_proj    & $30.7\%$ & $35.6\%$ & $\mathbf{48.5\%}$ \\
 & o\_proj    & $27.3\%$ & $28.3\%$ & $\mathbf{38.5\%}$ \\
 & gate\_proj & $24.6\%$ & $24.5\%$ & $\mathbf{31.1\%}$ \\
 & up\_proj   & $17.4\%$ & $19.4\%$ & $\mathbf{26.6\%}$ \\
 & down\_proj & $19.3\%$ & $20.5\%$ & $\mathbf{27.6\%}$ \\
\bottomrule
\end{tabular}
\end{table}

\begin{figure}[h]
\centering
\includegraphics[width=0.92\textwidth]{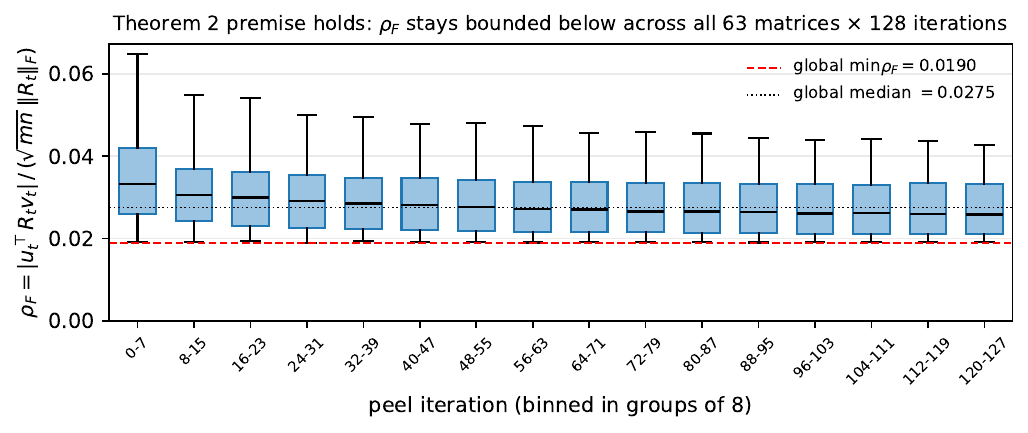}
\caption{Frobenius-relative capture rate over $128$ peel iterations
on Qwen3-4B projection matrices.}
\label{fig:empirical_rho}
\end{figure}

We also check the premise of the conditional cut-decay theorem empirically.
The premise of Theorem~\ref{thm:cut_decay_app} involves the cut norm $\|R_t\|_\blacksquare$, which is NP-hard to compute.
We instead measure the Frobenius analog $\rho_F(t) = |u_t^\top R_t v_t| / (\sqrt{mn}\,\|R_t\|_F)$.
The quantity is exactly the rate constant in $\|R_{t+1}\|_F^2 = \|R_t\|_F^2 (1 - \rho_F(t)^2)$ and by Cauchy--Schwarz it lower-bounds the cut-norm condition.
Figure~\ref{fig:empirical_rho} shows the distribution per iteration across $63$ Qwen3-4B matrices.
Over all $8064$ rank-one updates $\rho_F \geq 0.019$, and the median decays only mildly from $0.036$ at iteration $0$ to $0.026$ at iteration $127$. 
So in practice, the premise holds with $\rho_F$ bounded away from zero.

\subsection{Solver ablation, construction cost, and perplexity across ranks}
\label{app:solver_ablation}

Table~\ref{tab:solver_ablation} compares four ways to construct the $r{=}512$ low-rank correction on Llama-3-8B, using the setup of \S\ref{sec:exp_accuracy} and three sign seeds each. 
Random $\pm1$ sign vectors with the optimal joint coefficient refit give no improvement ($7.000$ against the uncorrected $6.998$ on WikiText-2), so all the gain comes from the learned signs. 
The other three constructions are greedy peeling, greedy peeling with the global refit,
and greedy peeling followed by $10$ alternating-minimization sweeps over the rank-one components (\emph{pm-altmin}), where each step removes one rank-one component, re-solves its sign alternation against the joint residual, and re-fits its
coefficient.
At this rank the three are within seed noise of one another.

\begin{table}[h]
\caption{Solver ablation on Llama-3-8B at $r{=}512$: perplexity (mean$\,\pm\,$std over three sign seeds) for four ways of constructing the low-rank component. Without a low-rank component (NVFP4 only), perplexity is $6.998$ on WikiText-2 and $7.680$ on C4.}
\label{tab:solver_ablation}
\centering
\small
\begin{tabular}{lrr}
\toprule
construction & WikiText-2 & C4 \\
\midrule
random signs $+$ joint coefficient refit & $7.000{\scriptstyle\,\pm 0.001}$ & $7.684{\scriptstyle\,\pm 0.008}$ \\
greedy learned signs                     & $6.842{\scriptstyle\,\pm 0.007}$ & $7.475{\scriptstyle\,\pm 0.018}$ \\
greedy learned signs $+$ global refit (Algorithm~\ref{alg:peel}) & $6.843{\scriptstyle\,\pm 0.017}$ & $7.480{\scriptstyle\,\pm 0.011}$ \\
$+$ pm-altmin, $10$ sweeps               & $6.838{\scriptstyle\,\pm 0.013}$ & $7.459{\scriptstyle\,\pm 0.014}$ \\
\bottomrule
\end{tabular}
\end{table}

\paragraph{Offline construction cost.}
Table~\ref{tab:ptq_cost} breaks the Llama-3-8B PTQ pass into its stages on one node. 
Building $512$ binary rank-one component greedily takes $86$\,s,
half the $170$\,s of the rank-$32$ SVD construction it replaces.
The global coefficient refit adds ${\sim}5$\,s, and residual NVFP4 quantization with bookkeeping takes under a second. 
Full alternating minimization is an order of magnitude more expensive and it is not part of the default construction.

\begin{table}[h]
\caption{PTQ time on Llama-3-8B in seconds on one node. ``Other'' is the NVFP4 quantization of the residual and bookkeeping. The PTQ total also includes collecting the calibration activations.}
\label{tab:ptq_cost}
\centering
\small
\begin{tabular}{lrrrr}
\toprule
method & solver & other & layer total & PTQ total \\
\midrule
SVDQuant, BF16 rank $32$ (SVD)       & $170.4$  & $2.5$ & $174.3$  & $222$ \\
greedy \method{}, $r{=}512$           & $86.3$   & $0.7$ & $90.3$   & $116$ \\
greedy $+$ global refit, $r{=}512$    & $91.3$   & $0.2$ & $92.5$   & $118$ \\
batched pm-altmin, $r{=}512$     & $213.9$  & $0.7$ & $215.9$  & $263$ \\
full pm-altmin, $r{=}512$             & $1070.2$ & $0.2$ & $1071.3$ & $1097$ \\
\bottomrule
\end{tabular}
\end{table}

\paragraph{Perplexity across correction ranks.}
Table~\ref{tab:ppl_rank_sweep} sweeps the rank on Llama-3-8B with setup of \S\ref{sec:exp_accuracy}. 
Perplexity improves monotonically with rank and
keeps improving past the matched-storage point, so $r{=}512$ is set by the byte budget rather than accuracy.
Ranks $r{=}1024$ and $r{=}2048$, at $2\times$ and $4\times$ the storage of SVDQuant $k{=}32$,  improve further, but the kernel overhead also grows with ranks
(Table~\ref{tab:binarysvd_kernel}).

\begin{table}[h]
\caption{\method{} perplexity versus rank on Llama-3-8B (C4-train
calibration, pm-altmin)}
\label{tab:ppl_rank_sweep}
\centering
\small
\begin{tabular}{lrr}
\toprule
 & WikiText-2 & C4 \\
\midrule
BF16                          & $6.266$ & $6.690$ \\
NVFP4 only                    & $6.998$ & $7.680$ \\
SVDQuant, BF16 rank $32$      & $6.919$ & $7.592$ \\
\method{} $r{=}32$            & $6.983$ & $7.656$ \\
\method{} $r{=}128$           & $6.926$ & $7.605$ \\
\method{} $r{=}256$           & $6.892$ & $7.549$ \\
\method{} $r{=}512$           & $6.835$ & $7.466$ \\
\method{} $r{=}1024$          & $6.749$ & $7.333$ \\
\method{} $r{=}2048$          & $6.627$ & $7.133$ \\
\bottomrule
\end{tabular}
\end{table}

\subsection{Composition with Rotated Backbones}
\label{app:rotation}

Rotation-based W4A4 methods rotate weights and activations so that outliers spread across channels. 
Table~\ref{tab:rotation_full} replaces SmoothQuant presmoothing with two full-width backbones on Llama-3-8B.
QuaRot~\cite{ashkboos2024quarotoutlierfree4bitinference} absorbs a $4096$-dimensional Hadamard rotation $R_1$ into the residual stream and adds a block-$2048$ rotation before the down projection.
SpinQuant~\cite{liu2025spinquantllmquantizationlearned} instead learns a global $R_1$, parametrized by the Cayley transform, initialized at the Hadamard matrix, and trained on the calibration cross-entropy of the quantized model with straight-through gradients through the FP4 fake-quant. 
Every other setting is unchanged, and the KV cache stays in BF16, so the KV rotation $R_2$ does not apply. 
\method{} improves both rotated backbones, and on each it is better than SVDQuant at the same storage. 
The FP4-only rows are worse than the presmoothed backbone, and the control without any preprocessing indicates that rotation itself hurts under NVFP4.
We attribute this to rotation spreading localized outliers across the $16$-element scaling blocks. 
The experiment therefore shows that \method{} is compatible with QuaRot and SpinQuant, not that rotation is the preferred NVFP4 preprocessing, and it uses one model and one seed.

\begin{table}[h]
\caption{Perplexity of Llama-3-8B with full-width rotated backbones (seed $42$, lower is better). SVDQuant uses BF16 rank $32$, and \method{} uses $r{=}512$ with the pm-altmin solver (Appendix~\ref{app:solver_ablation}).}
\label{tab:rotation_full}
\centering
\small
\begin{tabular}{lrr}
\toprule
backbone / correction & WikiText-2 & C4 \\
\midrule
no preprocessing, NVFP4 only            & $7.056$ & $7.751$ \\
presmoothing, NVFP4 only                & $6.998$ & $7.680$ \\
\midrule
QuaRot, NVFP4 only                      & $7.206$ & $7.981$ \\
QuaRot $+$ SVDQuant                     & $7.130$ & $7.904$ \\
QuaRot $+$ \method{}                    & $\mathbf{6.999}$ & $\mathbf{7.731}$ \\
\midrule
SpinQuant, NVFP4 only                   & $7.240$ & $8.035$ \\
SpinQuant $+$ SVDQuant                  & $7.158$ & $7.923$ \\
SpinQuant $+$ \method{}                 & $\mathbf{7.014}$ & $\mathbf{7.745}$ \\
\bottomrule
\end{tabular}
\end{table}

\subsection{Full Downstream Task Table}
\label{app:full_downstream}

Table~\ref{tab:downstream_full} reports the per-task PTQ accuracy on five standard PTQ tasks for three of the models in Table~\ref{tab:downstream_openllm}.
ARC-easy, ARC-challenge, PIQA, and OBQA report \texttt{acc\_norm}, and WinoGrande reports \texttt{acc}. 
All entries use the setup of\S\ref{sec:exp_accuracy}: C4-train calibration, SmoothQuant presmoothing $\alpha{=}0.5$, and the \texttt{mse}-based NVFP4 scale search.
\method{} $r{=}512$ has a higher five-task mean on Qwen3-4B and Qwen3-14B and is $0.3$ points lower on Llama-3-8B, where \method{} $r{=}128$ has the highest mean.

\begin{table}[h]
\caption{Standard PTQ downstream task breakdown. Higher is better.}
\label{tab:downstream_full}
\centering
\small
\begin{tabular}{llrcccccc}
\toprule
Model & Method & $b$ & ARC-e & ARC-c & PIQA & WinoG. & OBQA & mean \\
\midrule
\multirow{3}{*}{Qwen3-4B}
  & SVDQuant $k{=}32$    & 512 & 0.775 & 0.467 & \textbf{0.738} & 0.639 & 0.296 & 0.583 \\
  \cmidrule{2-9}
  & \method{} $r{=}128$  & 128 & \textbf{0.789} & 0.485 & 0.727 & 0.618 & \textbf{0.298} & 0.583 \\
  & \method{} $r{=}512$  & 512 & 0.782 & \textbf{0.486} & 0.724 & \textbf{0.652} & 0.286 & \textbf{0.586} \\
\midrule
\multirow{3}{*}{Qwen3-14B}
  & SVDQuant $k{=}32$    & 512 & \textbf{0.830} & 0.557 & 0.780 & \textbf{0.717} & 0.344 & 0.646 \\
  \cmidrule{2-9}
  & \method{} $r{=}128$  & 128 & \textbf{0.830} & 0.560 & \textbf{0.795} & 0.688 & 0.344 & 0.643 \\
  & \method{} $r{=}512$  & 512 & 0.828 & \textbf{0.583} & \textbf{0.795} & 0.715 & \textbf{0.348} & \textbf{0.654} \\
\midrule
\multirow{3}{*}{Llama-3-8B}
  & SVDQuant $k{=}32$    & 512 & 0.787 & 0.472 & 0.775 & 0.716 & 0.328 & 0.616 \\
  \cmidrule{2-9}
  & \method{} $r{=}128$  & 128 & \textbf{0.790} & \textbf{0.485} & \textbf{0.777} & \textbf{0.728} & \textbf{0.338} & \textbf{0.624} \\
  & \method{} $r{=}512$  & 512 & 0.767 & 0.469 & 0.774 & \textbf{0.728} & 0.326 & 0.613 \\
\bottomrule
\end{tabular}
\\\smallskip
\end{table}

\subsection{Full KL Table}
\label{app:full_kl}

Beyond perplexity and downstream accuracy, we measure the mean and 99th-percentile per-token Kullback--Leibler divergence $\mathrm{KL}(p_{\mathrm{BF16}}\,\|\,p_{\mathrm{quant}})$ between the BF16 reference model and each quantized model on $200$ held-out FineWeb samples (Table~\ref{tab:kl_full}). Table~\ref{tab:kl_full} covers three models at two storage budgets, and Table~\ref{tab:kl} adds Qwen3-32B and Qwen2.5-72B at $b{=}512$.
At $b{=}512$, \method{} has lower mean KL than SVDQuant on all three models (Qwen3-4B $0.074$ vs.\ $0.084$, Qwen3-14B $0.048$ vs.\ $0.052$, Llama-3-8B $0.076$ vs.\ $0.082$) and also lower $\text{KL}_{p99}$ on all three. 
At $b{=}128$, the results are mixed. \method{} is better on both metrics for Llama-3-8B, better at the 99th percentile but slightly worse on the mean for Qwen3-4B, and worse on both for Qwen3-14B. 

\begin{table}[h]
\caption{KL divergence to the BF16 model at $b{=}128$ and $b{=}512$ on held-out FineWeb samples. Lower is better.}
\label{tab:kl_full}
\centering
\small
\begin{tabular}{llrcccc}
\toprule
& & & \multicolumn{2}{c}{KL mean ($\downarrow$)}
    & \multicolumn{2}{c}{KL p99 ($\downarrow$)} \\
\cmidrule(lr){4-5}\cmidrule(lr){6-7}
Model & Method & $b$ & \method{} & SVDQ & \method{} & SVDQ \\
\midrule
\multirow{2}{*}{Qwen3-4B}
  & $r{=}128$ vs.\ $k{=}8$  & 128 & 0.0914 & \textbf{0.0901} & \textbf{0.1849} & 0.1965 \\
  & $r{=}512$ vs.\ $k{=}32$ & 512 & \textbf{0.0740} & 0.0835 & \textbf{0.1421} & 0.1649 \\
\midrule
\multirow{2}{*}{Qwen3-14B}
  & $r{=}128$ vs.\ $k{=}8$  & 128 & 0.0548 & \textbf{0.0541} & 0.2026 & \textbf{0.1618} \\
  & $r{=}512$ vs.\ $k{=}32$ & 512 & \textbf{0.0480} & 0.0517 & \textbf{0.1303} & 0.1671 \\
\midrule
\multirow{2}{*}{Llama-3-8B}
  & $r{=}128$ vs.\ $k{=}8$  & 128 & \textbf{0.0856} & 0.0881 & \textbf{0.3611} & 0.4789 \\
  & $r{=}512$ vs.\ $k{=}32$ & 512 & \textbf{0.0756} & 0.0824 & \textbf{0.3252} & 0.3409 \\
\bottomrule
\end{tabular}
\end{table}

\subsection{Fused-Kernel Implementation}
\label{app:high_rank_notes}

\paragraph{Implementation details.}
\label{app:kernel_details}
The kernel zero-pads the FP4 activation to the GEMM's full 128-row tile.
At small $M$, the block-scaled FP4 GEMM without padding is $23$--$36\%$ slower, because every CTA re-reads the same small $A$ tile at every $k$-step.
Padding is valid because the scale-factor layout is already padded to 128 rows and the padded output rows are discarded.
The correction is still computed only for the $M$ real rows, and the FP4-only baseline in \S\ref{sec:exp_kernel} uses the same padding.

At $K{=}N{=}5120$ the grid has $80$ GEMM CTAs, one per $64$-column output tile, and up to $68$ side CTAs on the remaining SMs.
The side CTAs compute $z=\alpha\odot(Vx)$ on CUDA cores directly from the packed signs of $V$.
Each value is written with an epoch tag, so a GEMM CTA can tell a value from the current launch from a stale one without clearing the buffer between launches.
In each GEMM CTA, the four epilogue warps are idle while the tensor cores run the main loop.
They wait for $z$ and expand it through the packed signs of $U$ for the tile's $64$ columns, using $16$-entry lookup tables in shared memory, one per group of four ranks and token row.
Each lookup replaces four sign-selected additions with one table read.
The result is added to the FP32 accumulators before the BF16 store, so the correction never makes a separate pass over the output.
When $M r$ is large ($M{=}16$ with $r{\geq}256$, and $M{=}32$ with $r{\geq}128$), all CTAs instead compute the correction at kernel entry and the GEMM CTAs start the GEMM afterwards.
This is still one launch, but the correction is no longer hidden under the GEMM.

\paragraph{Equal-rank control: binary versus BF16 factors.}
\label{app:equal_rank}
To separate the effect of the one-bit format from the effect of the fusion, we run a dense BF16 rank-$32$ correction, which has SVDQuant's factor format and the same storage as binary $r{=}512$, inside the same kernel as the binary branch.
The side CTAs compute $z$ from dense $V$ and the epilogue expands it through dense $U$.
At $M{=}1$ and $K{=}N{=}5120$, the dense rank-$32$ kernel takes $6.20\,\mu$s, the same as the binary $r{=}32$ kernel, against $5.84\,\mu$s for the padded FP4-only GEMM in the same run.
The same dense correction run unfused, as two BF16 GEMMs followed by the FP4 GEMM with a residual add, takes $17.64\,\mu$s.
At equal rank the two formats cost the same, so the one-bit format does not make the arithmetic faster.
Its advantage is the rank that a fixed storage budget buys: binary $r{=}512$ runs in $6.49\,\mu$s (Table~\ref{tab:binarysvd_kernel}) and improves accuracy over BF16 rank $32$ (Table~\ref{tab:ppl_b512}).

%% file: sections/a1_proofs.tex
\section{Proofs}
\label{app:proofs}

We include the formal statements and proofs for the signed-cut claims referenced in Section~\ref{sec:why_works}.

\subsection{Monotone Single-term Sign Alternation}
\label{app:proof_per_step}

\begin{theorem}[Monotone single-term sign alternation]
\label{thm:per_step_app}
Run Algorithm~\ref{alg:peel}'s inner loop from any sign initialization $v^{(0)} \in \{\pm 1\}^n$, alternate $u^{(t)} = \mathrm{sign}(R v^{(t-1)})$ and $v^{(t)} = \mathrm{sign}(R^\top u^{(t)})$ with deterministic tie-breaking, and set $\alpha^\star = (u^\top R v)/(mn)$ where $u, v$ are the final iterates. Then $u^\top Rv$ is non-decreasing after each  $u$- or $v$-update. 
The deflated residual $R' = R - \alpha^\star u v^\top$ satisfies
\[
\|R\|_F^2 - \|R'\|_F^2
\;=\; \frac{(u^\top R v)^2}{mn}.
\]
\end{theorem}

\paragraph{Monotonicity.} For fixed $u \in \{\pm 1\}^m$, the $v$-update $v \leftarrow \mathrm{sign}(R^\top u)$ maximizes $u^\top R v$ over $v \in \{\pm 1\}^n$ (sign-of-each-coordinate is a separable maximization). 
Symmetrically, the $u$-update maximizes $u^\top R v$ over $u \in \{\pm 1\}^m$. 
Thus each $u$- or $v$-update is non-decreasing in the single-term objective. With deterministic tie-breaking, the finite state space implies convergence to a fixed point or plateau.

\paragraph{Frobenius decrease.} For any $u \in \{\pm 1\}^m$, $v \in \{\pm 1\}^n$ and $R'  = R - \alpha u v^\top$, $\|R'\|_F^2 = \|R\|_F^2 - 2\alpha \,u^\top R v + \alpha^2 m n$. 
Minimizing in $\alpha$ gives $\alpha^\star = u^\top R v / (mn)$ and $\|R\|_F^2 - \|R'\|_F^2 = (u^\top R v)^2 / (mn)$. $\square$

\subsection{Conditional Signed-cut Decay}
\label{app:proof_decay}

\begin{theorem}[Conditional signed-cut decay]
\label{thm:cut_decay_app}
Let $W \in \mathbb{R}^{m \times n}$ and run a peeling procedure for $r$ steps. 
At step $t$, let $R_t$ be the residual and suppose the selected signs satisfy
\[
|u_t^\top R_t v_t| \;\geq\; \rho \|R_t\|_\blacksquare
\qquad\text{for some }\rho \in (0,1].
\]
With $\alpha_t = (u_t^\top R_t v_t)/(mn)$ and $R_{t+1} = R_t - \alpha_t u_t v_t^\top$, the residuals satisfy
\[
\min_{0 \leq t < r} \|R_t\|_\blacksquare^2
\;\leq\;
\frac{mn}{\rho^2 r}\,\|W\|_F^2.
\]
\end{theorem}

Let $\Phi_t = \|R_t\|_F^2$. The oracle condition and Theorem~\ref{thm:per_step}'s Frobenius identity give
\[
\Phi_t - \Phi_{t+1}
\;=\;
\frac{(u_t^\top R_t v_t)^2}{mn}
\;\geq\;
\frac{\rho^2\|R_t\|_\blacksquare^2}{mn}.
\]
Since $\Phi_0 = \|W\|_F^2$ and $\Phi_r \geq 0$, telescoping yields
\[
\sum_{t=0}^{r-1} \|R_t\|_\blacksquare^2
\;\leq\; \frac{mn}{\rho^2}\,\|W\|_F^2.
\]
At least one of the $r$ residuals therefore satisfies
$\|R_t\|_\blacksquare^2 \leq mn\|W\|_F^2/(\rho^2 r)$. This proves the stated minimum-over-iterations bound. No monotonicity of the final cut norm is required or claimed. $\square$

\subsection{Optimal Coefficient Refit}
\label{app:proof_alpha_refit}

\begin{theorem}[Optimal $\alpha$ refit]
\label{thm:alpha_refit_app}
Given fixed binary $\{u_i, v_i\}_{i=1}^r$, the closed-form solution $\boldsymbol{\alpha}^* = A^{-1}b$ with $A_{ij} = (u_i^\top u_j)(v_i^\top v_j)$ and $b_i = u_i^\top W v_i$ is the unique minimizer of $\Phi(\boldsymbol{\alpha}) = \|W - \sum_i \alpha_i u_i v_i^\top\|_F^2$, provided $A$ is invertible. 
The matrix $A$ is positive semi-definite and is strictly positive definite whenever the terms are linearly independent in $\mathbb{R}^{m \times n}$.
\end{theorem}

Let $a_i = \mathrm{vec}(u_i v_i^\top)$ and let $G = [a_1,\dots,a_r] \in \mathbb{R}^{mn \times r}$. Then
\[
\Phi(\boldsymbol{\alpha})
= \|\mathrm{vec}(W) - G\boldsymbol{\alpha}\|_2^2 .
\]
The normal equations are
\[
G^\top G \boldsymbol{\alpha} = G^\top \mathrm{vec}(W).
\]
Their entries are
\[
(G^\top G)_{ij}
= \langle u_i v_i^\top, u_j v_j^\top\rangle_F
= (u_i^\top u_j)(v_i^\top v_j)
\]
and
\[
(G^\top \mathrm{vec}(W))_i = u_i^\top W v_i .
\]
Thus $A = G^\top G$ and $b = G^\top \mathrm{vec}(W)$, giving $A\boldsymbol{\alpha}=b$. The Gram form makes $A$ positive semi-definite. If the terms are linearly independent, $G$ has full column rank and $A$ is positive definite, so the minimizer is unique.
With a numerical ridge, the implemented solve is $(A+\epsilon I)^{-1}b$. $\square$

%% file: sections/references.bib
@inproceedings{guo-etal-2024-fast,
    title = "Fast Matrix Multiplications for Lookup Table-Quantized {LLM}s",
    author = "Guo, Han  and
      Brandon, William  and
      Cholakov, Radostin  and
      Ragan-Kelley, Jonathan  and
      Xing, Eric P.  and
      Kim, Yoon",
    editor = "Al-Onaizan, Yaser  and
      Bansal, Mohit  and
      Chen, Yun-Nung",
    booktitle = "Findings of the Association for Computational Linguistics: EMNLP 2024",
    month = nov,
    year = "2024",
    address = "Miami, Florida, USA",
    publisher = "Association for Computational Linguistics",
    url = "https://aclanthology.org/2024.findings-emnlp.724/",
    doi = "10.18653/v1/2024.findings-emnlp.724",
    pages = "12419--12433"
}

@article{li2023loftq,
  title={Loftq: Lora-fine-tuning-aware quantization for large language models},
  author={Li, Yixiao and Yu, Yifan and Liang, Chen and He, Pengcheng and Karampatziakis, Nikos and Chen, Weizhu and Zhao, Tuo},
  journal={arXiv preprint arXiv:2310.08659},
  year={2023}
}

@misc{frantar2023gptqaccurateposttrainingquantization,
      title={GPTQ: Accurate Post-Training Quantization for Generative Pre-trained Transformers}, 
      author={Elias Frantar and Saleh Ashkboos and Torsten Hoefler and Dan Alistarh},
      year={2023},
      eprint={2210.17323},
      archivePrefix={arXiv},
      primaryClass={cs.LG},
      url={https://arxiv.org/abs/2210.17323}, 
}

@inproceedings{10.1145/1007352.1007371,
author = {Alon, Noga and Naor, Assaf},
title = {Approximating the cut-norm via Grothendieck's inequality},
year = {2004},
isbn = {1581138520},
publisher = {Association for Computing Machinery},
address = {New York, NY, USA},
url = {https://doi.org/10.1145/1007352.1007371},
doi = {10.1145/1007352.1007371},
booktitle = {Proceedings of the Thirty-Sixth Annual ACM Symposium on Theory of Computing},
pages = {72–80},
numpages = {9},
location = {Chicago, IL, USA},
series = {STOC '04}
}

@misc{yuan2025asvdactivationawaresingularvalue,
      title={ASVD: Activation-aware Singular Value Decomposition for Compressing Large Language Models}, 
      author={Zhihang Yuan and Yuzhang Shang and Yue Song and Dawei Yang and Qiang Wu and Yan Yan and Guangyu Sun},
      year={2025},
      eprint={2312.05821},
      archivePrefix={arXiv},
      primaryClass={cs.CL},
      url={https://arxiv.org/abs/2312.05821}, 
}

@misc{wang2025svdllmv2optimizingsingular,
      title={SVD-LLM V2: Optimizing Singular Value Truncation for Large Language Model Compression}, 
      author={Xin Wang and Samiul Alam and Zhongwei Wan and Hui Shen and Mi Zhang},
      year={2025},
      eprint={2503.12340},
      archivePrefix={arXiv},
      primaryClass={cs.CL},
      url={https://arxiv.org/abs/2503.12340}, 
}

@misc{saha2024compressinglargelanguagemodels,
      title={Compressing Large Language Models using Low Rank and Low Precision Decomposition}, 
      author={Rajarshi Saha and Naomi Sagan and Varun Srivastava and Andrea J. Goldsmith and Mert Pilanci},
      year={2024},
      eprint={2405.18886},
      archivePrefix={arXiv},
      primaryClass={cs.LG},
      url={https://arxiv.org/abs/2405.18886}, 
}

@misc{wang2025svdllmtruncationawaresingularvalue,
      title={SVD-LLM: Truncation-aware Singular Value Decomposition for Large Language Model Compression}, 
      author={Xin Wang and Yu Zheng and Zhongwei Wan and Mi Zhang},
      year={2025},
      eprint={2403.07378},
      archivePrefix={arXiv},
      primaryClass={cs.CL},
      url={https://arxiv.org/abs/2403.07378}, 
}

@misc{staats2025smallsingularvaluesmatter,
      title={Small Singular Values Matter: A Random Matrix Analysis of Transformer Models}, 
      author={Max Staats and Matthias Thamm and Bernd Rosenow},
      year={2025},
      eprint={2410.17770},
      archivePrefix={arXiv},
      primaryClass={cs.LG},
      url={https://arxiv.org/abs/2410.17770}, 
}

@misc{wang2023bitnetscaling1bittransformers,
      title={BitNet: Scaling 1-bit Transformers for Large Language Models}, 
      author={Hongyu Wang and Shuming Ma and Li Dong and Shaohan Huang and Huaijie Wang and Lingxiao Ma and Fan Yang and Ruiping Wang and Yi Wu and Furu Wei},
      year={2023},
      eprint={2310.11453},
      archivePrefix={arXiv},
      primaryClass={cs.CL},
      url={https://arxiv.org/abs/2310.11453}, 
}

@misc{shah2024flashattention3fastaccurateattention,
      title={FlashAttention-3: Fast and Accurate Attention with Asynchrony and Low-precision}, 
      author={Jay Shah and Ganesh Bikshandi and Ying Zhang and Vijay Thakkar and Pradeep Ramani and Tri Dao},
      year={2024},
      eprint={2407.08608},
      archivePrefix={arXiv},
      primaryClass={cs.LG},
      url={https://arxiv.org/abs/2407.08608}, 
}

@ARTICLE{9917312,
  author={Zhao, Han and Cui, Weihao and Chen, Quan and Guo, Minyi},
  journal={IEEE Transactions on Computers}, 
  title={ISPA: Exploiting Intra-SM Parallelism in GPUs via Fine-Grained Resource Management}, 
  year={2023},
  volume={72},
  number={5},
  pages={1473-1487},
  doi={10.1109/TC.2022.3214088}}

@misc{nvidiaIntroducingNVFP4,
    author = {{NVIDIA}},
	title = {{I}ntroducing {N}{V}{F}{P}4 for {E}fficient and {A}ccurate {L}ow-{P}recision {I}nference | {N}{V}{I}{D}{I}{A} {T}echnical {B}log --- developer.nvidia.com},
	howpublished = {\url{https://developer.nvidia.com/blog/introducing-nvfp4-for-efficient-and-accurate-low-precision-inference}},
    year = {2024},
	note = {[Accessed 05-05-2026]},
}

@Article{Frieze1999,
author={Frieze, Alan
and Kannan, Ravi},
title={Quick Approximation to Matrices and Applications},
journal={Combinatorica},
year={1999},
month={Feb},
day={01},
volume={19},
number={2},
pages={175-220},
issn={1439-6912},
doi={10.1007/s004930050052},
url={https://doi.org/10.1007/s004930050052}
}

@misc{zhang2026sparsebitnet158bitllmsnaturally,
      title={Sparse-BitNet: 1.58-bit LLMs are Naturally Friendly to Semi-Structured Sparsity}, 
      author={Di Zhang and Xun Wu and Shaohan Huang and Yudong Wang and Hanyong Shao and Yingbo Hao and Zewen Chi and Li Dong and Ting Song and Yan Xia and Zhifang Sui and Furu Wei},
      year={2026},
      eprint={2603.05168},
      archivePrefix={arXiv},
      primaryClass={cs.CL},
      url={https://arxiv.org/abs/2603.05168}, 
}

@misc{zadouri2026flashattention4algorithmkernelpipelining,
      title={FlashAttention-4: Algorithm and Kernel Pipelining Co-Design for Asymmetric Hardware Scaling}, 
      author={Ted Zadouri and Markus Hoehnerbach and Jay Shah and Timmy Liu and Vijay Thakkar and Tri Dao},
      year={2026},
      eprint={2603.05451},
      archivePrefix={arXiv},
      primaryClass={cs.CL},
      url={https://arxiv.org/abs/2603.05451}, 
}

@misc{guo2024lqloralowrankplusquantized,
      title={LQ-LoRA: Low-rank Plus Quantized Matrix Decomposition for Efficient Language Model Finetuning}, 
      author={Han Guo and Philip Greengard and Eric P. Xing and Yoon Kim},
      year={2024},
      eprint={2311.12023},
      archivePrefix={arXiv},
      primaryClass={cs.CL},
      url={https://arxiv.org/abs/2311.12023}, 
}

@misc{liu2025spinquantllmquantizationlearned,
      title={SpinQuant: LLM quantization with learned rotations}, 
      author={Zechun Liu and Changsheng Zhao and Igor Fedorov and Bilge Soran and Dhruv Choudhary and Raghuraman Krishnamoorthi and Vikas Chandra and Yuandong Tian and Tijmen Blankevoort},
      year={2025},
      eprint={2405.16406},
      archivePrefix={arXiv},
      primaryClass={cs.LG},
      url={https://arxiv.org/abs/2405.16406}, 
}

@misc{ashkboos2024quarotoutlierfree4bitinference,
      title={QuaRot: Outlier-Free 4-Bit Inference in Rotated LLMs}, 
      author={Saleh Ashkboos and Amirkeivan Mohtashami and Maximilian L. Croci and Bo Li and Pashmina Cameron and Martin Jaggi and Dan Alistarh and Torsten Hoefler and James Hensman},
      year={2024},
      eprint={2404.00456},
      archivePrefix={arXiv},
      primaryClass={cs.LG},
      url={https://arxiv.org/abs/2404.00456}, 
}

@misc{xiao2024smoothquantaccurateefficientposttraining,
      title={SmoothQuant: Accurate and Efficient Post-Training Quantization for Large Language Models}, 
      author={Guangxuan Xiao and Ji Lin and Mickael Seznec and Hao Wu and Julien Demouth and Song Han},
      year={2024},
      eprint={2211.10438},
      archivePrefix={arXiv},
      primaryClass={cs.CL},
      url={https://arxiv.org/abs/2211.10438}, 
}

@misc{lin2026awqactivationawareweightquantization,
      title={AWQ: Activation-aware Weight Quantization for LLM Compression and Acceleration}, 
      author={Ji Lin and Jiaming Tang and Haotian Tang and Shang Yang and Wei-Ming Chen and Wei-Chen Wang and Guangxuan Xiao and Xingyu Dang and Chuang Gan and Song Han},
      year={2026},
      eprint={2306.00978},
      archivePrefix={arXiv},
      primaryClass={cs.CL},
      url={https://arxiv.org/abs/2306.00978}, 
}

@misc{egiazarian2026bridginggappromiseperformance,
      title={Bridging the Gap Between Promise and Performance for Microscaling FP4 Quantization}, 
      author={Vage Egiazarian and Roberto L. Castro and Denis Kuznedelev and Andrei Panferov and Eldar Kurtic and Shubhra Pandit and Alexandre Marques and Mark Kurtz and Saleh Ashkboos and Torsten Hoefler and Dan Alistarh},
      year={2026},
      eprint={2509.23202},
      archivePrefix={arXiv},
      primaryClass={cs.LG},
      url={https://arxiv.org/abs/2509.23202}, 
}

@misc{cook2025sixaccuratenvfp4quantization,
      title={Four Over Six: More Accurate NVFP4 Quantization with Adaptive Block Scaling},
      author={Jack Cook and Junxian Guo and Guangxuan Xiao and Yujun Lin and Song Han},
      year={2025},
      eprint={2512.02010},
      archivePrefix={arXiv},
      primaryClass={cs.CL},
      url={https://arxiv.org/abs/2512.02010},
}

@inproceedings{10.1145/3710848.3710871,
author = {Frantar, Elias and Castro, Roberto L. and Chen, Jiale and Hoefler, Torsten and Alistarh, Dan},
title = {MARLIN: Mixed-Precision Auto-Regressive Parallel Inference on Large Language Models},
year = {2025},
isbn = {9798400714436},
publisher = {Association for Computing Machinery},
address = {New York, NY, USA},
url = {https://doi.org/10.1145/3710848.3710871},
doi = {10.1145/3710848.3710871},
booktitle = {Proceedings of the 30th ACM SIGPLAN Annual Symposium on Principles and Practice of Parallel Programming},
pages = {239–251},
numpages = {13},
location = {Las Vegas, NV, USA},
series = {PPoPP '25}
}

@inproceedings{
lin2025qservewakv,
title={{QS}erve:W4A8{KV}4 Quantization and System Co-design for Efficient {LLM} Serving},
author={Yujun Lin and Haotian Tang and Shang Yang and Zhekai Zhang and Guangxuan Xiao and Chuang Gan and Song Han},
booktitle={Eighth Conference on Machine Learning and Systems},
year={2025},
url={https://openreview.net/forum?id=1FfmStySS1}
}

@inproceedings{
li2025svdquant,
title={{SVDQ}uant: Absorbing Outliers by Low-Rank Component for 4-Bit Diffusion Models},
author={Muyang Li and Yujun Lin and Zhekai Zhang and Tianle Cai and Junxian Guo and Xiuyu Li and Enze Xie and Chenlin Meng and Jun-Yan Zhu and Song Han},
booktitle={The Thirteenth International Conference on Learning Representations},
year={2025},
url={https://openreview.net/forum?id=vWR3KuiQur}
}

@misc{riasanovsky2024efficient1bittensorapproximations,
      title={Efficient $1$-bit tensor approximations}, 
      author={Alex W. Neal Riasanovsky and Sarah El Kazdadi},
      year={2024},
      eprint={2410.01799},
      archivePrefix={arXiv},
      primaryClass={math.CO},
      url={https://arxiv.org/abs/2410.01799}, 
}
